# Evaluation of Momentum Diverse Input Iterative Fast Gradient Sign Method (M-DI2-FGSM) Based Attack Method on MCS 2018 Adversarial Attacks on Black Box Face Recognition System

**Md Ashraful Alam Milton**[1]
[1]Department of Computer Engineering, Autonomous University of Barcelona, Spain

**Abstract.** The convolutional neural network is the crucial tool for the recent success of deep learning based methods on various computer vision tasks like classification, segmentation, and detection. Convolutional neural networks achieved state-of-the-art performance in these tasks and every day pushing the limit of computer vision and AI. However, adversarial attack on computer vision systems is threatening their application in the real life and in safety-critical applications. Necessarily, Finding adversarial examples are important to detect susceptible models to attack and take safeguard measures to overcome the adversarial attacks. In this regard, MCS 2018 Adversarial Attacks on Black Box Face Recognition challenge aims to facilitate the research of finding new adversarial attack techniques and their effectiveness in generating adversarial examples. In this challenge, the attack's nature is targeted-attack on the black-box neural network where we have no knowledge about black-block's inner structure. The attacker must modify a set of five images of a single person so that the neural network miss-classify them as target image which is a set of five images of another person. In this competition, we applied Momentum Diverse Input Iterative Fast Gradient Sign Method (M-DI2-FGSM) to make an adversarial attack on black-box face recognition system. We tested our method on MCS 2018 Adversarial Attacks on Black Box Face Recognition challenge and found competitive result. Our solution got validation score 1.404 which better than baseline score 1.407 and stood 14 place among 132 teams in the leader-board. Further improvement can be achieved by finding improved feature extraction from source image, carefully chosen hyper-parameters, finding improved substitute model of the black-box and better optimization method. The Code is available for download in the following github link: https://github.com/miltonbd/mcs_2018_adversarial_attack

## 1 Introduction

After the last AI winter, since 2012 deep learning based neural networks are state-of-the-art technology for computer vision and achieved an unprecedented result on various vision tasks, including image classification [1,7-9], object detection [10-13] and semantic segmentation [14-16]. However, due CNN's internal structure and way of work, they are can be compromised by adversarial images with small modifications which is impossible to differentiate by the human eye. Deep neural networks (DNNs) can be compromised by their inability to tackle adversarial examples [4, 5], which carefully devised by adding small, human-imperceptible noises to legitimate examples as mentioned in figure 1. This attack is successful in both scenario of targeted and non-targeted attack. In targeted where the final in-accurate output set by the attacker and in no-targeted attack where the model's output can be any of the inaccurate predictions. Arguably, the time has come to take safeguard measures and incorporate built-in defenses against adversarial attack in the real-world machine learning system like the autonomous vehicle, drone vision, face recognition etc. systems.

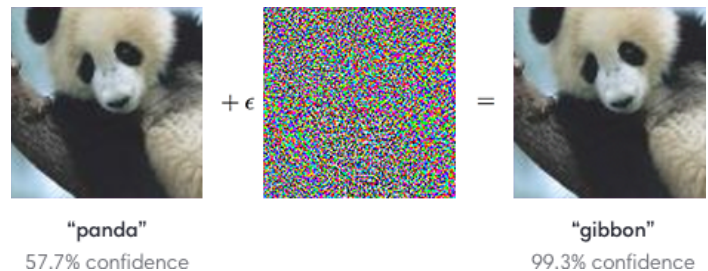

Fig 1. Adversarial example of the panda with added perturbation.

Images generated by adversarial attack can be employed to a different domain like image segmentation[24] and in reinforcement learning[25]. Several studies focus on understanding the insufficiency of current training algorithms [6,25,26,32] against adversarial examples. Many important methods [3,5,20,26,27,31] have been developed recently to find adversarial examples. Adversarial examples can be a severe security threat to practical computer vision applications. In particular, an adversarial example[4] that was designed to be miss-classified by a model A is often also miss-classified by a model B. This adversarial example's transferability property means that it is possible to generate adversarial examples and perform a miss-classification attack on a machine learning system without access to the underlying model. Such attacks can be done in the real world [29,30]. Data augmentation [7-9] has been proved to be a viable solution to prevent networks from overfitting during the training process. Clearly, a series of transformation like resizing, cropping and rotating, are applied to the images to enhance the training set, make robust model and thus prevent adversarial examples based attack. As a result, the trained networks will be robust unknown input and will generalize well. Also, image augmentation[20,31] can defend against adversarial examples under certain situations, which proves that adversarial examples cannot generalize well under different transformations. To this extent, we explored the Diverse Input Iterative Fast Gradient Sign Method (DI-2-FGSM) based attack on MCS 2018 Adversarial Attacks on Black Box Face Recognition to improve the transferability of adversarial examples. At each iteration, we diverse the inputs by applying augmentation using imgaug[2] library. This way the model has always some modified source image which reduces overfitting. This method performed better than the baseline method.

## 2 Background And Related work

In this section, we will review the background information and related works about the adversarial attack. The main goal of the adversarial attack is to fabricate new image by adding carefully controlled noise to the original image in such a way that the changes are almost undetectable to the human eye.

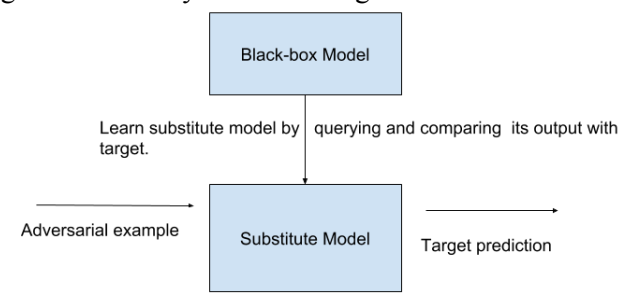

Fig 2. Black-box and its substitute model.

In the white box attack, the intruder has access to the model parameters and in the black box attack the intruder has no access to the model parameters. Thus, for the black box attack, a mechanism, to replicate the black box network's internal functionality is needed. The black box and substitute model relation is shown in figure 2. The goal of non-targeted attacks is to compel the model to miss-classify the adversarial image and in the targeted attack, the goal is to miss-classify the source image into a fixed wrong class. Most prominent attacks are gradient-based attacks where the attacker modifies the source image by observing the flow of gradient and other techniques. In this way, the attackers modify the image in the direction of the gradient of the loss function with respect to the input image. One-shot attacks and iterative attacks are popular in gradient-based attacks. In one-shot attack, where the attacker takes a single step in the direction of the gradient, and in the iterative attack where instead of a single step, several steps are used to adversarial exmaples. Several methods[22,33] have been developed to tackle the adversarial attacks. Both the adversarial attack and defense are being improved by ongoing research.

## 3  Attack Methods

In the recent years, extensive research brought about a variety of gradient based adversarial attack methods. The evolution of gradient-based attack are in following sections.

### 3.1  Fast Gradient Sign Method (FGSM)

FGSM[5] is the first member in this attack family, which finds the adversarial perturbations in the direction of the loss gradient and the adversarial example image generation equation can be expressed as:

$$X^{\text{adv}} = X + \epsilon \cdot sign(\nabla_X L(X, y^{\text{true}}; \theta)). \qquad \ldots\ldots\ldots\ldots\ldots(1)$$

### 3.2  Iterative Fast Gradient Sign Method (I-FGSM)

Kurakin et al. [27] extended FGSM to an iterative version, which can be expressed as:

$$X_0^{\text{adv}} = X,$$

$$X_{n+1}^{\text{adv}} = \text{Clip}_X^\epsilon \left\{ X_n^{adv} + \alpha \cdot sign(\nabla_X L(X_n^{adv}, y^{\text{true}}; \theta)) \right\} \qquad \ldots\ldots\ldots\ldots\ldots(2)$$

the original image X, n is the iteration number and α is the step size.

### 3.3  Momentum Iterative Fast Gradient Sign Method (MI-FGSM)

MI-FGSM was proposed to integrate the momentum term into the attack process to stabilize update directions and escape from poor local maxima. The updating procedure is similar to I-FGSM,

$$g_{n+1} = \mu \cdot g_n + \frac{\nabla_X L(X_n^{adv}, y^{\text{true}}; \theta)}{||\nabla_X L(X_n^{adv}, y^{\text{true}}; \theta)||_1},$$

$$X_{n+1}^{\text{adv}} = \text{Clip}_X^\epsilon \left\{ X_n^{adv} + \alpha \cdot sign(g_{n+1}) \right\}, \qquad \ldots\ldots\ldots\ldots\ldots(3)$$

Where μ is the decay factor of the momentum term and $g_n$ is the accumulated gradient at iteration n.

### 3.4  Diverse Input Iterative Fast Gradient Sign Method (DI2-FGSM)

In diverse input, various changes are applied to source image like random crop, gaussian blur with Sigma 0.5, contrast normalization, affine transformation with scaling, transformation, rotate and shear.

$$X_{n+1}^{\text{adv}} = \text{Clip}_X^\epsilon \left\{ X_n^{adv} + \alpha \cdot sign(\nabla_X L(T(X_n^{adv}; p), y^{\text{true}}; \theta)) \right\} \quad \ldots\ldots\ldots\ldots\ldots\ldots(4)$$

$$T(X_n^{adv}; p) = \begin{cases} T(X_n^{adv}), & \text{with probability } p \\ X_n^{adv}, & \text{with probability } 1-p \end{cases} \quad \ldots\ldots\ldots\ldots\ldots\ldots(5)$$

Momentum and diverse inputs work differently to prevent overfitting phenomenon. Combination both known as Momentum Diverse Inputs Iterative Fast Gradient Sign Method (M-DI2-FGSM) has good potential as a good adversarial attack method. The overall updating procedure of M-DI2 -FGSM is similar to MI-FGSM :

$$g_{n+1} = \mu \cdot g_n + \frac{\nabla_X L(T(X_n^{adv}; p), y^{\text{true}}; \theta)}{||\nabla_X L(T(X_n^{adv}; p), y^{\text{true}}; \theta)||_1}. \quad \ldots\ldots\ldots\ldots\ldots\ldots(6)$$

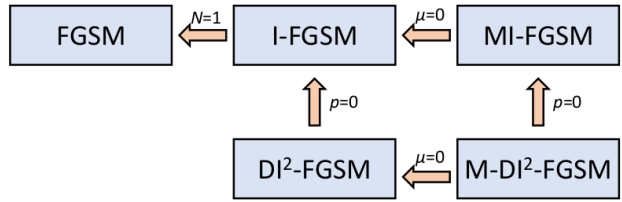

**Fig 3. The relationship among fast gradient step method.**

Figure 3 shows the relationship among gradient step methods. If the transformation[36] probability is p = 0, M-DI 2 -FGSM degrades to MI-FGSM, and DI-2 -FGSM degrades to I-FGSM; If the decay factor μ = 0, M-DI 2 -FGSM degrades to DI 2 -FGSM, and MI-FGSM degrades to I-FGSM, If the total iteration number N = 1, I-FGSM degrades to FGSM. In all above equations positive sign before the α indicates that the attack is non-targeted and for the targeted attacks the α sign will be negative.

## 4 Challenge Details

Faces are an important natural way to recognize people. It is the extension of general face detection to provide recognition of an individual. Recent deep learning and computer vision technology enables to scale up face recognition to correctly recognize millions of identity. Modern methods of face recognition easily surpass human performance while relying on machine learning and neural networks based techniques. This type of face recognition system will enable modern bio-metric system and other real-world applications. However, such face recognition system can be vulnerable to attacks aiming to fabricate the network's final output. Seemingly, arbitrary changes to the network output can be produced by small and well-designed modifications of the network input, as known under adversarial examples.

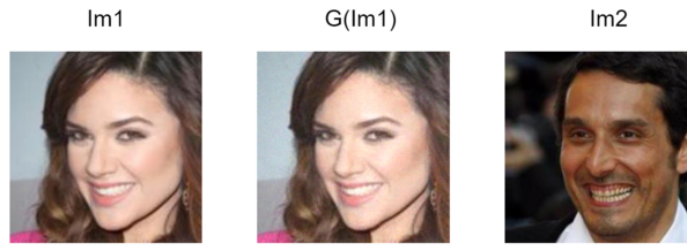

**Fig 4. MCS 2018 Adversarial Attacks on Black Box Face Recognition.**

The strategy for adversarial attack on face recognition system is given in figure 4. Here, Im1 is the original source image. G(Im1) is the adversarial image with perturbation crafted by attacker which will be miss-classified by face recognition system as Im2.

## 5 Proposed Methodology

### 5.1 Make Substitute Model

In order to guess the internal parameters and simulate the attack, we need a substitute model[28] black box. We used densenet based methods to make a substitute classification model by feeding the output of the model to our system as input and using our known labels of those inputs. We used 1M face recognition data provided by the organizer to train the classifier. MEAN = [0.485, 0.456, 0.406] STD = [0.229, 0.224, 0.225] used for normalizing the input data.

### 5.2 Fine Tuning

As the amounts of training examples is insufficient for training a deep convolutional network from scratch, so we used the imagenet pre-trained DenseNet model to initialize the network parameters, freeze the initial layers and fine-tuned the last several layers. We kept the weight of all layers except FC and output layer was frozen in first few iterations of training due to high unstable gradient flow. After few iterations of training, we unfreeze the last few layers and adjust their weight by backpropagation. Fine-tuning helps to train easily and prevents it from overfitting. Obviously, the fine-tuned network can lead to a better convergence.

### 5.3 Pre-Processing

First the images is center cropped from 224 to 112. Transforms. The source image pixels are normalized using MEAN [0.485, 0.456, 0.406] and STD = [0.229, 0.224, 0.225].

### 5.4 Diverse Input

we are given with 5 source images and 5 target images. We keep the target images same without augmentation and to satisfy the diverse input feature of the algorithm we augmented the 5 source face images to 20. Some augmented images are mentioned in figure 5.

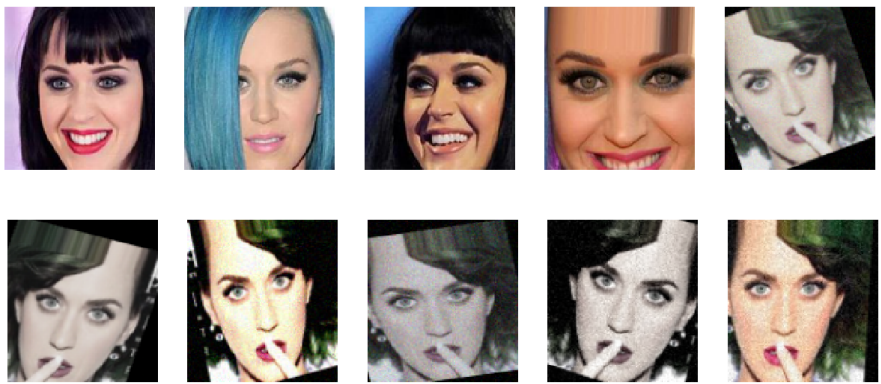

**Fig 5. Augmented images as diverse input.**

Data augmentation methods used were random crop, Gaussian blur with Sigma 0.5, contrast normalization, affine transformation with scaling, transformation, rotate and shear. We also applied grayscale conversion for some images. All augmentations were done using imgaug[1]. The input images with probability p went through the diverse input transformation and 1-p were fed to the model as usual.

### 5.5 Attack Details

We run the attack for maximum 60 epochs. For some images, we were having too much or too little gradient propagation and sometime it was found that after all iteration the SSIM is still 1 and another time after 1 iteration, it went below 0.95. In both cases, we adjusted the epsilon with either multiplying or dividing the initial eps with factor 2.

## 6 Evaluation

To evaluate the model performance, we fed the 1000 images to substitute model and generate the adversarial example. We also extracted a 1000*512 descriptor to be uploaded to the evaluation server as per directed by the organizer. For the evaluation of the result, we have to upload the generated adversarial images and 1000*512 face descriptor in a single zip file. The Structural Similarity Index (SSIM) was used for measuring original and generated image difference and that must be below above .95 threshold.

## 7 Implementation Details

For implementation and simulate the adversarial attack, PyTorch[23] deep learning framework was chosen. We used 2 1080TI GPU based pc with CUDA 9 and Ubuntu 16.04 OS. This computing power was not enough to get the result quickly, hence lowered the chance to ace the competition leaderboard.

## 8 Results and Discussion

Our Momentum Diverse Input Iterative Fast Gradient Sign Method (M-DI2-FGSM) achieved 1.404 in the final leaderboard which is better than baseline value of 1.407. We secured 14[th] position in the final leaderboard. Due to the late participation and the limited deadline, we did not get to the very top in the competition leaderboard and we believe there is ample opportunity to extend the attack method with more improved feature extraction from source image, carefully chosen hyper-parameters, and finding improved substitute model of the black-box. In addition to that, a bigger dataset with improved image augmentation reduces the risk of overfitting on the black-box model and thus, substitute model mimics the black-box network well. Moreover, performing additional regularization tweaks and fine-tuning of hyper-parameters may improve model's robustness. Along with densenet to make substitute model of black-box, other network models like SENet[17], NASNet[18], PNASNet[19] might be explored.

## 9 Conclusion

In this paper, we demonstrated that Momentum Diverse Input Iterative Fast Gradient Sign Method (M-DI2-FGSM) can be effectively applied to adversarial attack on the face recognition system. This could elicit a great opportunity to test the face recognition system against this robust adversarial attack and pave the way to take measures against adversarial attack. Moreover, Adversarial training, input sanitization are the important measures to make sure that the face recognition system has already taken care of the adversarial example of input images.